\documentclass[a4paper,twoside]{article}
\pdfoutput=1

\usepackage{epsfig}
\usepackage{subcaption}
\usepackage{calc}
\usepackage{amssymb}
\usepackage{amstext}
\usepackage{amsmath}
\usepackage{amsthm}
\usepackage{multicol}
\usepackage{pslatex}
\usepackage{apalike}
\usepackage{algorithm2e}
\usepackage[bottom]{footmisc}

\usepackage[T1]{fontenc}

\usepackage[pdftex, 
            pdfauthor={Vincent A. Cicirello},
            pdftitle={Open Source Evolutionary Computation with Chips-n-Salsa},
            pdfsubject={Cicirello, V.A. (2024). Open Source Evolutionary Computation with Chips-n-Salsa.  In Proceedings of the 16th International Joint Conference on Computational Intelligence (IJCCI 2024), pages 330-337. https://doi.org/10.5220/0013040600003837},
            pdfkeywords={Evolutionary Algorithms, Evolutionary Computation, Genetic Algorithms, Java, Open Source, Parallel, Self-Adaptive}]{hyperref}
			
\usepackage{breakurl}

\usepackage{fancyhdr}

\fancypagestyle{specialfooter}{%
  \fancyhf{}
  
  \fancyfoot[L]{Accepted version. Cite published version as:\\Cicirello, V.A. (2024). Open Source Evolutionary Computation with Chips-n-Salsa.  In {\it Proceedings of the 16th International Joint Conference on Computational Intelligence (IJCCI 2024)}, pages 330-337. \url{https://doi.org/10.5220/0013040600003837}}
}

\usepackage{SCITEPRESS}     

\SetKwProg{Interface}{public interface}{ \{}{\}}
\SetKwProg{Class}{public class}{ \{}{\}}
\SetKwFunction{MutationOperator}{MutationOperator}
\SetKwFunction{CrossoverOperator}{CrossoverOperator}
\SetKwFunction{Splittable}{Splittable}
\SetKwFunction{Copyable}{Copyable}
\SetKwFunction{SelectionOperator}{SelectionOperator}
\SetKwFunction{IntegerCostOptimizationProblem}{IntegerCostOptimizationProblem}
\SetKwFunction{OptimizationProblem}{OptimizationProblem}
\SetKwFunction{Configurator}{Configurator}
\SetKwFunction{Problem}{Problem}
\SetKwFunction{FitnessFunction}{FitnessFunction}
\SetKwFunction{FitnessFunctionDouble}{FitnessFunction.Double}
\SetKwFunction{FitnessFunctionInteger}{FitnessFunction.Integer}
\SetAlgorithmName{Listing}{listing}{List of Listings}

\begin{document}

\title{Open Source Evolutionary Computation with Chips-n-Salsa}
\thispagestyle{specialfooter}

\author{\authorname{Vincent A. Cicirello \orcidAuthor{0000-0003-1072-8559}}
\affiliation{Computer Science, School of Business, Stockton University, 101 Vera King Farris Dr, Galloway, NJ, USA}
\email{vincent.cicirello@stockton.edu}
}


\abstract{When it was first introduced, the Chips-n-Salsa Java library provided 
stochastic local search and related algorithms, with a focus on self-adaptation
and parallel execution. For the past four years, we expanded its scope to include 
evolutionary computation. This paper concerns the evolutionary algorithms that 
Chips-n-Salsa now provides, which includes multiple evolutionary models, common 
problem representations, a wide range of mutation and crossover operators, 
and a variety of benchmark problems. Well-defined Java interfaces enable easily 
integrating custom representations and evolutionary operators, as well as defining 
optimization problems. Chips-n-Salsa's evolutionary algorithms include implementations 
with adaptive mutation and crossover rates, as well as both sequential and parallel 
execution. Source code is maintained on GitHub, and immutable artifacts are regularly 
published to the Maven Central Repository to enable easily importing into projects for 
reproducible builds. Effective development processes such as test-driven development, 
as well as a variety of static analysis tools help ensure code quality.}

\onecolumn \maketitle \normalsize \setcounter{footnote}{0} \vfill

\section{\uppercase{Introduction}}
\label{sec:introduction}

Evolutionary computation refers to the family of problem solving frameworks
that are inspired by models of natural evolution and genetics. Most forms of 
evolutionary computation are population-based, maintaining a population
of many candidate solutions to a problem, which evolve over many generations
using operators that mimic evolutionary processes such as mutation and 
recombination. Evolutionary computation is often used to solve a wide variety
of problems, including in software engineering~\cite{Sobania2023,Arcuri2021,Petke2018},
computer vision and image processing~\cite{Wan2023,Bi2023}, neural network 
construction~\cite{Zhou2021}, engineering design~\cite{Tayarani2015}, production 
scheduling~\cite{Branke2016}, finance~\cite{Ponsich2013}, graph 
theory~\cite{Pizzuti2018}, feature selection~\cite{Xue2016}, data 
mining~\cite{Mukhopadhyay2014}, multiobjective optimization~\cite{Liang2023}, 
dynamic optimization problems~\cite{Yazdani2021}, among many others.

Previously, we introduced Chips-n-Salsa~\cite{cicirello2020joss}, an open 
source Java library for stochastic local search and related algorithms. At the time, 
Chips-n-Salsa version 1.3.0 did not include evolutionary computation. Instead 
it focused on a variety of other metaheuristics, such as hill 
climbing~\cite{Hoos2018,Selman2006,PrugelBennett2004}, simulated 
annealing~\cite{Delahaye2019}, and stochastic sampling search 
algorithms~\cite{Grasas2017,Cicirello2005,Gomes1998,Bresina1996,Langley1992}. 
Chips-n-Salsa supports self-adaptive search, such as adaptive annealing
schedules~\cite{Cicirello2021,Hubin2019,Stefankovic2009} for simulated annealing, 
and adaptive restart schedules~\cite{Cicirello2017,Luby1993} for multistart search. 
Chips-n-Salsa is also designed to easily support parallel search, 
as well as to enable defining hybrids of multiple metaheuristics. The algorithm 
implementations are highly customizable such as in choice of search operators, 
including providing well-defined interfaces to enable the option to integrate 
custom components with library components. Chips-n-Salsa currently requires 
Java 17 at minimum. Source code is maintained on GitHub, and API documentation 
and other information available on the web. Immutable artifacts of every release, 
including pre-compiled jar of the library and jars of the source and documentation, 
are regularly published to the Maven Central Repository to enable easily importing 
into software development projects, as well as to ensure reproducible builds. See 
Table~\ref{tab:url} for library URLs.

\begin{table}[t]
\caption{URLs for Chips-n-Salsa.}\label{tab:url}
\centering
\addtolength{\tabcolsep}{-3pt}
\begin{tabular}{@{}lp{168pt}@{}}\hline
Source & \url{https://github.com/cicirello/Chips-n-Salsa} \\
Website & \url{https://chips-n-salsa.cicirello.org/} \\
Maven & \url{https://central.sonatype.com/artifact/org.cicirello/chips-n-salsa/} \\
Docs & \url{https://chips-n-salsa.cicirello.org/api/} \\
Examples & \url{https://github.com/cicirello/chips-n-salsa-examples} \\
\hline
\end{tabular}
\addtolength{\tabcolsep}{3pt}
\end{table}

During the four years since that previous publication, there have been six major releases
of Chips-n-Salsa, whose scope we expanded to include evolutionary computation. 
This paper focuses on version 7.0.0, and more specifically on the genetic algorithms (GA) 
and other evolutionary algorithms (EA) that the library now provides. Chips-n-Salsa 
supports multiple forms of EA, such as the classic generational EA as well as the 
$(\mu + \lambda)$-EA, and also includes many crossover and mutation operators for 
different representations, such as for optimizing bit-vectors, permutations, and vectors of 
reals and integers. Additionally, to complement Chips-n-Salsa's emphasis on self-adaptation,
it includes an adaptive EA where the crossover and mutation rates evolve during the search.
We have also added many common benchmarking problems to the library. The objectives of doing
so include serving as a research testbed, and supporting reproducible empirical 
evolutionary computation research. Reproducible research~\cite{NAP25303} is crucial in
all fields of science and engineering.

In the remainder of this paper, we cover the EA functionality of Chips-n-Salsa as of version
7.0.0. We begin with a brief discussion of related evolutionary computation libraries in 
Section~\ref{sec:related}. Then, in Section~\ref{sec:functionality}, we present the EAs of 
Chips-n-Salsa, along with a discussion of key features of the implementations, and 
a discussion of the crossover and mutation operators available for each supported representation. 
We also summarize the benchmark problems implemented within the library. Our objective isn't only
to enable reproducible research, but also to provide a production ready open source toolkit for
use by practitioners. Thus, we employ high-quality development practices, which we summarize in 
Section~\ref{sec:practices}. We wrap up in Section~\ref{sec:conclusion}.

\section{\uppercase{Related Work}}\label{sec:related}

There are several other open source libraries available for evolutionary
computation~\cite{Jenetics,Tarkowski2023,Dios2022,Izzo2020,Scott2019,Bell2019,Gijsbers2019,Detorakis2019,Simson2019}.

Many focus on a specific form of evolutionary computation. For example, 
LGP~\cite{Simson2019} implements genetic programming (GP) in Kotlin. 
DCGP~\cite{Izzo2020} is also a GP library, but in C++ with a Python
interface. Metaheuristics~\cite{Dios2022} is a Julia package consisting 
of several metaheuristics for both single-objective and multi-objective 
optimization. CEGO~\cite{Bell2019} is a differential evolution C++ library 
with a Python wrapper. Quil\"{e}~\cite{Tarkowski2023} and 
GAIM~\cite{Detorakis2019} are both C++ libraries for GAs,
with GAIM focused specifically on multi-population island models. 
ECJ~\cite{Scott2019} and Jenetics~\cite{Jenetics} are both Java libraries 
supporting multiple forms of evolutionary computation.

Chips-n-Salsa differs from the existing libraries in a few ways. First,
it supports both evolutionary computation as well as other metaheuristics.
Due to this, it is straightforward to create hybrids of multiple techniques.
Second, observing that much of the runtime of an EA is spent generating
random numbers, we have highly optimized the randomness of Chips-n-Salsa
utilizing the $\rho\mu$ library~\cite{cicirello2022joss}. Third, Chips-n-Salsa
includes more built-in support for evolving permutations compared to other
libraries, with a comprehensive collection of evolutionary permutation
operators~\cite{cicirello2023ecta}. Additionally, Chips-n-Salsa is designed
to enable parallel execution.

\section{\uppercase{Core Functionality}}\label{sec:functionality}

This Section provides an overview of the core evolutionary computation
functionality of the Chips-n-Salsa library. It is organized 
into subsections covering the EA features and characteristics 
(Section~\ref{sec:ea}), the evolutionary operators provided by the 
library (Section~\ref{sec:ops}), and the available benchmark problems 
(Section~\ref{sec:problems}).

\subsection{Evolutionary Algorithms}\label{sec:ea}

\textbf{Evolutionary Models:} Chips-n-Salsa provides implementations of both
generational EAs, where each generation involves replacing the current population
with an offspring population produced via application of the crossover and
mutation operators, as well as steady-state EAs, where a small number of children 
are generated at a time, replacing a small number of members of the population.
For example, the $(\mu + \lambda)$-EA maintains a population of size $\mu$, generates
a small number of children $\lambda$, and keeps the $\mu$ best of the combination.
The simplest form of steady-state EA is the $(\mu + 1)$-EA that generates a single
offspring at a time. The library also includes the special case of the $(1 + 1)$-EA.
Chips-n-Salsa additionally supports the special case of a mutation-only EA.
The generational EAs include an option for elitism, where a small number of the
best population members survive without undergoing crossover or mutation.

\textbf{Representations:} Chips-n-Salsa provides several built-in representations,
including efficient bit-vectors suitable for a GA, as well as vectors of integers 
and reals, such as for real-valued function optimization. Since many combinatorial 
optimization problems concern searching for an optimal ordering, Chips-n-Salsa 
supports evolving permutations utilizing an efficient permutation 
representation~\cite{cicirello2018joss}. Section~\ref{sec:ops} focuses on the 
evolutionary operators available for each of these representations.

\textbf{Customizable:} If the built-in representations are unsuitable for a problem, 
Chips-n-Salsa supports custom representations via Java generic types and by 
implementing a Java interface \Copyable as seen in Listing~\ref{alg:copyable},
which enables the library to create identical copies of population members
as needed in a representation-independent manner. Defining evolutionary operators 
for the new representation is accomplished by implementing
the \MutationOperator and \CrossoverOperator interfaces shown in Listing~\ref{alg:operators}.
Those same interfaces also enable implementing custom crossover and mutation operators for the 
built-in representations.

\begin{algorithm}[t]
\caption{Interface to define custom representation.}\label{alg:copyable}
\DontPrintSemicolon
\SetAlgoLined
\SetKwFunction{Copy}{copy}%
\Interface{\Copyable<T \BlockMarkersSty{extends} \Copyable<T>{}>}{
  T \Copy{T $c$};
}
\end{algorithm}

\begin{algorithm}[t]
\caption{Interfaces for evolutionary operators.}\label{alg:operators}
\DontPrintSemicolon
\SetAlgoLined
\SetKwFunction{Mutate}{mutate}%
\SetKwFunction{Cross}{cross}%
\SetKwFunction{Split}{split}%
\Interface{\MutationOperator<T> \BlockMarkersSty{extends} \Splittable<\MutationOperator<T>{}>}{
  \BlockMarkersSty{void} \Mutate{T $c$};
}
\;
\Interface{\CrossoverOperator<T> \BlockMarkersSty{extends} \Splittable<\CrossoverOperator<T>{}>}{
  \BlockMarkersSty{void} \Cross{T $c1$, T $c2$};
}
\;
\Interface{\Splittable<T \BlockMarkersSty{extends} \Splittable<T>{}>}{
  T \Split{};
}
\end{algorithm}

\textbf{Adaptive:} In addition to the more common case of control parameters that
remain constant throughout a run, Chips-n-Salsa includes an implementation of an
adaptive EA that encodes the crossover and mutation rates as part of each population 
member, using Gaussian mutation to evolve these during the search~\cite{Hinterding1995}.

\textbf{Parallel:} All metaheuristics in the library implement a set of Java 
interfaces, enabling the EA implementations to utilize the library's existing 
parallel architecture for multi-populations to accelerate runtime on multicore systems.
One of these interfaces is the \Splittable interface shown earlier in 
Listing~\ref{alg:operators}, which enables the library's parallel architecture to 
replicate the functionality of operators, etc when spawning threads.

\textbf{Hybridization:} Hybrids of EA with other search algorithms are common. For example,
a memetic algorithm combines an EA with local search~\cite{Neri2012}. Creating such hybrids
in Chips-n-Salsa is straightforward, facilitated by its plug-and-play design.

\textbf{Configurable and optimized randomness:} The runtime of an EA or GA can be
significantly impacted by the choice of pseudorandom number generator 
(PRNG)~\cite{Nesmachnow2015}, as well as the choice of algorithm for generating 
values from specific distributions (e.g., uniform subject to bounds, Gaussian, Cauchy, etc). 
For this reason, we carefully optimized all random behavior within Chips-n-Salsa
utilizing the enhanced random functionality of the $\rho\mu$ library~\cite{cicirello2022joss}. 
Additionally, by default, Chips-n-Salsa uses a carefully chosen, efficient PRNG,
but also provides the option to configure the PRNG through its \Configurator
class whose interface is shown in Listing~\ref{alg:random}.

\begin{algorithm}[t]
\caption{Random generator configuration.}\label{alg:random}
\DontPrintSemicolon
\SetAlgoLined
\SetKwFunction{ConfigureRandomGenerator}{configureRandomGenerator}%
\SetKwFunction{RandomGenerator}{RandomGenerator.SplittableGenerator}%
\SetKwData{Seed}{seed}
\SetKwData{R}{r}
\Class{\Configurator}{
  \BlockMarkersSty{public} \BlockMarkersSty{static} \BlockMarkersSty{void} \ConfigureRandomGenerator{\BlockMarkersSty{long} \Seed}; \;
  \BlockMarkersSty{public} \BlockMarkersSty{static} \BlockMarkersSty{void} \ConfigureRandomGenerator{\\ \RandomGenerator \R}; 
}
\end{algorithm}

\subsection{Evolutionary Operators}\label{sec:ops}

\textbf{Selection Operators:} The selection operators~\cite{Mitchell1998} include 
fitness proportionate selection (i.e., weighted roulette wheel), truncation selection, 
tournament selection, stochastic universal sampling (SUS), linear rank selection, 
exponential rank selection, Boltzmann selection, and random selection. There are 
two forms of each of linear rank, exponential rank, and Boltzmann selection: one that 
operates like roulette wheel and one that operates like SUS. There are also options to 
transform fitness values relative to the population fitness during selection, such as 
with sigma scaling or by shifting the fitness scale. An interface, \SelectionOperator, 
is provided to enable defining custom selection operators (see Listing~\ref{alg:selection}).

\begin{algorithm}[t]
\caption{Interface for defining selection operators.}\label{alg:selection}
\DontPrintSemicolon
\SetAlgoLined
\SetKwFunction{Init}{init}%
\SetKwFunction{Select}{select}%
\SetKwData{Generations}{generations}%
\SetKwData{Selected}{selected}%
\SetKwData{Fitnesses}{fitnesses}%
\Interface{\SelectionOperator \BlockMarkersSty{extends} \Splittable<\SelectionOperator>}{
  \BlockMarkersSty{default} \BlockMarkersSty{void} \Init{\BlockMarkersSty{int} \Generations}; \;
  \BlockMarkersSty{void} \Select{ PopulationFitnessVector.Double \Fitnesses, \BlockMarkersSty{int[]} \Selected}; \;
  \BlockMarkersSty{void} \Select{ PopulationFitnessVector.Integer \Fitnesses, \BlockMarkersSty{int[]} \Selected}; \;
}
\end{algorithm}

\textbf{Evolutionary Operators for Bit Vectors:} Chips-n-Salsa supports the common
bit-flip mutation for mutating bit vectors, and all of the common crossover operators, such
as single-point, two-point, $k$-point, and uniform crossover operators.

\textbf{Evolutionary Operators for Integer Vectors and Real Vectors:} For vectors of
integers and reals, the library includes the obvious extensions of single-point, 
two-point, $k$-point, and uniform crossover. For mutating real vectors, the library 
includes Gaussian mutation~\cite{Hinterding1995}, Cauchy mutation~\cite{Szu1987}, and
uniform mutation. For integer vectors, it includes a uniform mutation (e.g., that
adds a uniform random value from $[-w, w]$ to components of the vector), as well as a
random value change mutation that replaces a value with a different random value from
its domain.

\textbf{Evolutionary Operators for Permutations:} Many evolutionary operators exist for
evolving permutations~\cite{cicirello2023ecta}. Chips-n-Salsa provides an 
extensive collection of crossover and mutation operators for permutations. 
Mutation operators include all of the common permutation mutation 
operators~\cite{cicirello2023ecta,Serpell2010,eiben03,Valenzuela2001} 
as well as a few less common ones. The mutation operators for permutations include: 
swap, adjacent swap, insertion, reversal, scramble, uniform scramble, 
cycle mutation~\cite{cicirello2022applsci}, 
3opt~\cite{Lin1965}, block move, block swap, as well as 
window-limited mutation operators~\cite{cicirello2014bict}. 
Crossover operators include: cycle crossover~\cite{Oliver1987}, position based 
crossover~\cite{Barecke2007}, order crossover~\cite{Davis1985}, order 
crossover 2~\cite{Syswerda1991}, non-wrapping order crossover~\cite{cicirello2006gecco},
uniform order based crossover~\cite{Syswerda1991}, partially matched 
crossover~\cite{Goldberg1985}, uniform partially matched 
crossover~\cite{cicirello2000gecco}, edge recombination~\cite{Whitley1989}, enhanced 
edge recombination~\cite{Starkweather1991}, precedence preservative 
crossover~\cite{Bierwirth1996}, and uniform precedence preservative 
crossover~\cite{Bierwirth1996}.

\textbf{Hybrid Operators:} The library includes support for utilizing hybrid crossover
operators and hybrid mutation operators. Specifically, it includes classes to integrate
multiple crossover operators or mutation operators, such that an operator is chosen randomly 
from a specified set during each application. The random choice can be weighted (e.g., to choose
one operator with higher probability than another) or it can be a uniform random choice.

\subsection{Benchmark Problems}\label{sec:problems}

To solve optimization problems using the EAs of the library, you either implement 
the fitness function directly via a Java interface, or if it is a minimization 
problem, you can implement its cost function via a Java interface, and then use 
one of the library's classes for mapping a cost function (e.g., minimization) to 
a fitness function (e.g., maximization). Listing~\ref{alg:fitness} shows the interfaces
for defining fitness functions; and Listing~\ref{alg:problems} shows the
interfaces for defining optimization problems (\OptimizationProblem for real-valued
costs and \IntegerCostOptimizationProblem for integer-valued costs). The library
treats optimization functions and fitness functions as distinct concepts. 

\begin{algorithm}[t]
\caption{Interfaces to define fitness functions.}\label{alg:fitness}
\DontPrintSemicolon
\SetAlgoLined
\SetKwFunction{GetProblem}{getProblem}%
\SetKwFunction{Fitness}{fitness}%
\SetKwData{Candidate}{candidate}%
\Interface{\FitnessFunctionDouble<T \BlockMarkersSty{extends} \Copyable<T>{}> \BlockMarkersSty{extends} \FitnessFunction<T>}{
  \BlockMarkersSty{double} \Fitness{T \Candidate}; \;
}
\;
\Interface{\FitnessFunctionInteger<T \BlockMarkersSty{extends} \Copyable<T>{}> \BlockMarkersSty{extends} \FitnessFunction<T>}{
  \BlockMarkersSty{int} \Fitness{T \Candidate}; \;
}
\;
\Interface{\FitnessFunction<T \BlockMarkersSty{extends} \Copyable<T>{}>}{
  \Problem<T> \GetProblem{};
}
\end{algorithm}

\begin{algorithm}[t]
\caption{Interfaces to define problems.}\label{alg:problems}
\DontPrintSemicolon
\SetAlgoLined
\SetKwFunction{Cost}{cost}%
\SetKwFunction{CostAsDouble}{costAsDouble}%
\SetKwFunction{GetSolutionCostPair}{getSolutionCostPair}%
\SetKwFunction{IsMinCost}{isMinCost}%
\SetKwFunction{MinCost}{minCost}%
\SetKwFunction{Value}{value}%
\SetKwData{Candidate}{candidate}%
\SetKwData{CostVar}{cost}%
\Interface{\OptimizationProblem<T \BlockMarkersSty{extends} \Copyable<T>{}> \BlockMarkersSty{extends} \Problem<T>}{
  \BlockMarkersSty{double} \Cost{T \Candidate}; \;
  \BlockMarkersSty{double} \Value{T \Candidate}; \;
  \BlockMarkersSty{default} \BlockMarkersSty{double} \CostAsDouble{T \Candidate}; \;
  \BlockMarkersSty{default} SolutionCostPair<T> \GetSolutionCostPair{T \Candidate}; \;
  \BlockMarkersSty{default} \BlockMarkersSty{boolean} \IsMinCost{\BlockMarkersSty{double} \CostVar}; \;
  \BlockMarkersSty{default} \BlockMarkersSty{double} \MinCost{}; \;
}
\;
\Interface{\IntegerCostOptimizationProblem<T \BlockMarkersSty{extends} \Copyable<T>{}> \BlockMarkersSty{extends} \Problem<T>}{
  \BlockMarkersSty{int} \Cost{T \Candidate}; \;
  \BlockMarkersSty{int} \Value{T \Candidate}; \;
  \BlockMarkersSty{default} \BlockMarkersSty{double} \CostAsDouble{T \Candidate}; \;
  \BlockMarkersSty{default} SolutionCostPair<T> \GetSolutionCostPair{T \Candidate}; \;
  \BlockMarkersSty{default} \BlockMarkersSty{boolean} \IsMinCost{\BlockMarkersSty{int} \CostVar}; \;
  \BlockMarkersSty{default} \BlockMarkersSty{int} \MinCost{}; \;
}
\;
\Interface{\Problem<T \BlockMarkersSty{extends} \Copyable<T>{}>}{
  \BlockMarkersSty{double} \CostAsDouble{T \Candidate}; \;
  SolutionCostPair<T> \GetSolutionCostPair{T \Candidate}; \;
}
\end{algorithm}

Chips-n-Salsa includes many common benchmarking problems. For example,
it includes implementations of all of Ackley's~\cite{Ackley1985,Ackley1987} classic 
problems for bit vectors, such as one-max, two-max, trap, porcupine, plateaus, and mix,
as well as various royal roads problems~\cite{Mitchell1992,Holland1993,Jones1994}.
It includes some real-valued function optimization problems~\cite{Forrester2008,Gramacy2012}.
It also includes the permutation in a haystack~\cite{cicirello2016evc}, a benchmarking 
problem for permutations, as well as many NP-Hard combinatorial optimization 
problems~\cite{Garey1979}, such as the traveling salesperson, bin packing, largest 
common subgraph, quadratic assignment, and many scheduling problems.

\section{\uppercase{Development Practices}}\label{sec:practices}

In developing Chips-n-Salsa, we utilize best practices in software engineering,
such as test-driven development, integrating static analysis tools into the build
pipeline, and continuous integration and continuous deployment (CI/CD). Frequent 
public releases are deployed to artifact repositories (e.g., Maven Central and GitHub 
Packages) to ease potential integration into other open source projects. The immutable 
nature of the artifacts published to the Maven Central Repository help ensure builds of 
projects that depend upon Chips-n-Salsa are reproducible.

A few of the quality control methods used in developing Chips-n-Salsa are
as follows:

\textbf{Unit testing:} Following test-driven development practices, unit tests are written
for all components in the JUnit framework. 

\textbf{Regression testing:} All new and existing test cases must pass before changes
are accepted, to ensure that new functionality does not introduce bugs into existing
components.

\textbf{Test coverage:} We use the test coverage tool, JaCoCo~\cite{jacoco}, to compute both 
C0 coverage (instructions coverage) and C1 coverage (branches coverage) for all builds.

\textbf{Pull-request checks:} Chips-n-Salsa is developed openly on GitHub. GitHub's
built-in CI/CD framework, GitHub Actions~\cite{actions}, is used during pull-requests 
to verify that all test cases pass, and to run the test coverage analysis.

\textbf{Static analysis:} The build pipeline runs several static analysis tools to
automatically detect error prone code, including: CodeQL~\cite{codeql}, 
RefactorFirst~\cite{refactor-first}, SpotBugs~\cite{spotbugs}, FindSecBugs~\cite{findsecbugs}, 
and Snyk~\cite{snyk}.

\textbf{Code style:} For consistency, we use Google Java Style~\cite{google-style}, 
and automatically reformat to this style using a Maven plugin~\cite{spotify}.

\textbf{Dependency management:} We keep dependencies up to date with dependabot~\cite{dependabot}. 

\textbf{Documentation site:} The project website, among other things, contains the
Javadoc formatted documentation of the library (URL in Table~\ref{tab:url}), which is 
automatically updated during the release process.

\textbf{Example code:} In addition to the repository of the library itself, an additional
GitHub repository (URL in Table~\ref{tab:url}) provides a collection of detailed examples 
demonstrating how to use Chips-n-Salsa in projects.

Chips-n-Salsa is licensed via the GNU 
General Public License v3.0~\cite{GPL3}, and it welcomes contributions from the open 
source community, with well-defined guidelines for contributors.

\section{\uppercase{Conclusion}}\label{sec:conclusion}

Evolutionary computation is widely used in many fields. 
Chips-n-Salsa provides a well-engineered open source framework for EA and related 
metaheuristics, including a comprehensive set of evolutionary operators for common
representations like bit vectors, permutations, and vectors of integers
or reals, as well as implementations of many common benchmarking problems. It thus
can serve as an ideal framework for empirical research. For example, if a researcher 
implements their new EA using the library, they benefit from an easy way to compare
their approach to many others as well as on many problems. Deploying immutable software 
artifacts of each version of the library to Maven Central better enables reproducible 
research~\cite{NAP25303}, as future runs of experiments can use the exact versions of 
components as the original runs.

\bibliographystyle{apalike}
{\small
\bibliography{openevo}}

\begin{thebibliography}{}

\bibitem[Ackley, 1985]{Ackley1985}
Ackley, D.~H. (1985).
\newblock A connectionist algorithm for genetic search.
\newblock In {\em ICGA}, pages 121--135.

\bibitem[Ackley, 1987]{Ackley1987}
Ackley, D.~H. (1987).
\newblock An empirical study of bit vector function optimization.
\newblock In {\em Genetic Algorithms and Simulated Annealing}, pages 170--204.
  Morgan Kaufmann.

\bibitem[Arcuri et~al., 2021]{Arcuri2021}
Arcuri, A., Galeotti, J.~P., Marculescu, B., and Zhang, M. (2021).
\newblock Evomaster: A search-based system test generation tool.
\newblock {\em JOSS}, 6(57):2153.

\bibitem[Arteau, 2024]{findsecbugs}
Arteau, P. (2024).
\newblock Find security bugs: The spotbugs plugin for security audits of java
  web applications.
\newblock \url{https://find-sec-bugs.github.io/}.

\bibitem[Barecke and Detyniecki, 2007]{Barecke2007}
Barecke, T. and Detyniecki, M. (2007).
\newblock Memetic algorithms for inexact graph matching.
\newblock In {\em IEEE CEC}, pages 4238--4245.

\bibitem[Bell, 2019]{Bell2019}
Bell, I.~H. (2019).
\newblock {CEGO}: C++11 evolutionary global optimization.
\newblock {\em JOSS}, 4(36):1147.

\bibitem[Bethancourt, 2024]{refactor-first}
Bethancourt, J. (2024).
\newblock {RefactorFirst}.
\newblock \url{https://github.com/refactorfirst/RefactorFirst}.

\bibitem[Bi et~al., 2023]{Bi2023}
Bi, Y., Xue, B., Mesejo, P., Cagnoni, S., and Zhang, M. (2023).
\newblock A survey on evolutionary computation for computer vision and image
  analysis: Past, present, and future trends.
\newblock {\em IEEE TEVC}, 27(1):5--25.

\bibitem[Bierwirth et~al., 1996]{Bierwirth1996}
Bierwirth, C., Mattfeld, D.~C., and Kopfer, H. (1996).
\newblock On permutation representations for scheduling problems.
\newblock In {\em PPSN}, pages 310--318.

\bibitem[Branke et~al., 2016]{Branke2016}
Branke, J., Nguyen, S., Pickardt, C.~W., and Zhang, M. (2016).
\newblock Automated design of production scheduling heuristics: A review.
\newblock {\em IEEE TEVC}, 20(1):110--124.

\bibitem[Bresina, 1996]{Bresina1996}
Bresina, J.~L. (1996).
\newblock Heuristic-biased stochastic sampling.
\newblock In {\em AAAI}, pages 271--278.

\bibitem[Cicirello, 2006]{cicirello2006gecco}
Cicirello, V.~A. (2006).
\newblock Non-wrapping order crossover: An order preserving crossover operator
  that respects absolute position.
\newblock In {\em Proceedings of the Genetic and Evolutionary Computation
  Conference}, pages 1125--1131.

\bibitem[Cicirello, 2014]{cicirello2014bict}
Cicirello, V.~A. (2014).
\newblock On the effects of window-limits on the distance profiles of
  permutation neighborhood operators.
\newblock In {\em International Conference on Bio-inspired Information and
  Communication Technologies}, pages 28--35.

\bibitem[Cicirello, 2016]{cicirello2016evc}
Cicirello, V.~A. (2016).
\newblock The permutation in a haystack problem and the calculus of search
  landscapes.
\newblock {\em IEEE Transactions on Evolutionary Computation}, 20(3):434--446.

\bibitem[Cicirello, 2017]{Cicirello2017}
Cicirello, V.~A. (2017).
\newblock Variable annealing length and parallelism in simulated annealing.
\newblock In {\em International Symposium on Combinatorial Search}, pages
  2--10.

\bibitem[Cicirello, 2018]{cicirello2018joss}
Cicirello, V.~A. (2018).
\newblock {JavaPermutationTools}: A java library of permutation distance
  metrics.
\newblock {\em Journal of Open Source Software}, 3(31):950.

\bibitem[Cicirello, 2020]{cicirello2020joss}
Cicirello, V.~A. (2020).
\newblock {Chips-n-Salsa}: A java library of customizable, hybridizable,
  iterative, parallel, stochastic, and self-adaptive local search algorithms.
\newblock {\em Journal of Open Source Software}, 5(52):2448.

\bibitem[Cicirello, 2021]{Cicirello2021}
Cicirello, V.~A. (2021).
\newblock Self-tuning lam annealing: Learning hyperparameters while problem
  solving.
\newblock {\em Applied Sciences}, 11(21):9828.

\bibitem[Cicirello, 2022a]{cicirello2022applsci}
Cicirello, V.~A. (2022a).
\newblock Cycle mutation: Evolving permutations via cycle induction.
\newblock {\em Applied Sciences}, 12(11):5506.

\bibitem[Cicirello, 2022b]{cicirello2022joss}
Cicirello, V.~A. (2022b).
\newblock $\rho\mu$: A java library of randomization enhancements and other
  math utilities.
\newblock {\em Journal of Open Source Software}, 7(76):4663.

\bibitem[Cicirello, 2023]{cicirello2023ecta}
Cicirello, V.~A. (2023).
\newblock A survey and analysis of evolutionary operators for permutations.
\newblock In {\em Proceedings of the 15th International Joint Conference on
  Computational Intelligence}, pages 288--299.

\bibitem[Cicirello and Smith, 2000]{cicirello2000gecco}
Cicirello, V.~A. and Smith, S.~F. (2000).
\newblock Modeling ga performance for control parameter optimization.
\newblock In {\em Proceedings of the Genetic and Evolutionary Computation
  Conference}, pages 235--242.

\bibitem[Cicirello and Smith, 2005]{Cicirello2005}
Cicirello, V.~A. and Smith, S.~F. (2005).
\newblock Enhancing stochastic search performance by value-biased randomization
  of heuristics.
\newblock {\em Journal of Heuristics}, 11(1):5--34.

\bibitem[Davis, 1985]{Davis1985}
Davis, L. (1985).
\newblock Applying adaptive algorithms to epistatic domains.
\newblock In {\em IJCAI}, pages 162--164.

\bibitem[de~Dios and Mezura-Montes, 2022]{Dios2022}
de~Dios, J.-A.~M. and Mezura-Montes, E. (2022).
\newblock Metaheuristics: A julia package for single- and multi-objective
  optimization.
\newblock {\em JOSS}, 7(78):4723.

\bibitem[Delahaye et~al., 2019]{Delahaye2019}
Delahaye, D., Chaimatanan, S., and Mongeau, M. (2019).
\newblock Simulated annealing: From basics to applications.
\newblock In {\em Handbook of Metaheuristics}, pages 1--35. Springer.

\bibitem[Detorakis and Burton, 2019]{Detorakis2019}
Detorakis, G. and Burton, A. (2019).
\newblock Gaim: A c++ library for genetic algorithms and island models.
\newblock {\em JOSS}, 4(44):1839.

\bibitem[Eiben and Smith, 2003]{eiben03}
Eiben, A.~E. and Smith, J.~E. (2003).
\newblock {\em Introduction to Evolutionary Computing}.
\newblock Springer, Berlin, Germany.

\bibitem[Forrester et~al., 2008]{Forrester2008}
Forrester, A. I.~J., Sóbester, A., and Keane, A.~J. (2008).
\newblock Appendix: Example problems.
\newblock In {\em Engineering Design via Surrogate Modelling: A Practical
  Guide}, pages 195--203. Wiley.

\bibitem[{Free Software Foundation}, 2007]{GPL3}
{Free Software Foundation} (2007).
\newblock Gnu general public license.
\newblock \url{https://www.gnu.org/licenses/gpl-3.0.en.html}.

\bibitem[Garey and Johnson, 1979]{Garey1979}
Garey, M.~R. and Johnson, D.~S. (1979).
\newblock {\em Computers and Intractability: A Guide to the Theory of
  NP-Completeness}.
\newblock W. H. Freeman \& Co., USA.

\bibitem[Gijsbers and Vanschoren, 2019]{Gijsbers2019}
Gijsbers, P. and Vanschoren, J. (2019).
\newblock Gama: Genetic automated machine learning assistant.
\newblock {\em JOSS}, 4(33):1132.

\bibitem[{GitHub}, 2024a]{codeql}
{GitHub} (2024a).
\newblock {CodeQL}.
\newblock \url{https://codeql.github.com/}.

\bibitem[{GitHub}, 2024b]{dependabot}
{GitHub} (2024b).
\newblock Dependabot.
\newblock \url{https://github.com/dependabot/dependabot-core}.

\bibitem[{GitHub}, 2024c]{actions}
{GitHub} (2024c).
\newblock Github actions.
\newblock \url{https://github.com/features/actions}.

\bibitem[Goldberg and Lingle, 1985]{Goldberg1985}
Goldberg, D.~E. and Lingle, R. (1985).
\newblock Alleles, loci, and the traveling salesman problem.
\newblock In {\em ICGA}, pages 154--159.

\bibitem[Gomes et~al., 1998]{Gomes1998}
Gomes, C.~P., Selman, B., and Kautz, H. (1998).
\newblock Boosting combinatorial search through randomization.
\newblock In {\em AAAI}, pages 431--437.

\bibitem[Google, 2024]{google-style}
Google (2024).
\newblock Google java style guide.
\newblock \url{https://google.github.io/styleguide/}.

\bibitem[Gramacy and Lee, 2012]{Gramacy2012}
Gramacy, R.~B. and Lee, H. K.~H. (2012).
\newblock Cases for the nugget in modeling computer experiments.
\newblock {\em Statistics and Computing}, 22(3):713--722.

\bibitem[Grasas et~al., 2017]{Grasas2017}
Grasas, A., Juan, A.~A., Faulin, J., {de Armas}, J., and Ramalhinho, H. (2017).
\newblock Biased randomization of heuristics using skewed probability
  distributions: A survey and some applications.
\newblock {\em CAIE}, 110:216--228.

\bibitem[Hinterding, 1995]{Hinterding1995}
Hinterding, R. (1995).
\newblock Gaussian mutation and self-adaption for numeric genetic algorithms.
\newblock In {\em CEC}, pages 384--389.

\bibitem[Hoffmann et~al., 2024]{jacoco}
Hoffmann, M.~R., Janiczak, B., and Mandrikov, E. (2024).
\newblock {JaCoCo} java code coverage library.
\newblock \url{https://www.jacoco.org/jacoco/}.

\bibitem[Holland, 1993]{Holland1993}
Holland, J. (1993).
\newblock Royal road functions.
\newblock {\em Internet Genetic Algorithms Digest}, 7(22).

\bibitem[Hoos and St{\"{u}}tzle, 2018]{Hoos2018}
Hoos, H.~H. and St{\"{u}}tzle, T. (2018).
\newblock Stochastic local search.
\newblock In {\em Handbook of Approximation Algorithms and Metaheuristics
  Methologies and Traditional Applications}, chapter~17. Chapman and Hall/CRC,
  2nd edition.

\bibitem[Hubin, 2019]{Hubin2019}
Hubin, A. (2019).
\newblock An adaptive simulated annealing em algorithm for inference on
  non-homogeneous hidden markov models.
\newblock In {\em AIIPCC}, pages 1--9.

\bibitem[Izzo and Biscani, 2020]{Izzo2020}
Izzo, D. and Biscani, F. (2020).
\newblock dcgp: Differentiable cartesian genetic programming made easy.
\newblock {\em JOSS}, 5(51):2290.

\bibitem[Jenetics, 2024]{Jenetics}
Jenetics (2024).
\newblock Jenetics: Genetic algorithm, genetic programming, evolutionary
  algorithm, and multi-objective optimization.
\newblock \url{https://jenetics.io/}.

\bibitem[Jones, 1994]{Jones1994}
Jones, T. (1994).
\newblock A description of holland's royal road function.
\newblock {\em Evolutionary Computation}, 2(4):409--415.

\bibitem[Langley, 1992]{Langley1992}
Langley, P. (1992).
\newblock Systematic and nonsystematic search strategies.
\newblock In {\em AIPS}, pages 145--152.

\bibitem[Liang et~al., 2023]{Liang2023}
Liang, J., Ban, X., Yu, K., Qu, B., Qiao, K., Yue, C., Chen, K., and Tan, K.~C.
  (2023).
\newblock A survey on evolutionary constrained multiobjective optimization.
\newblock {\em IEEE TEVC}, 27(2):201--221.

\bibitem[Lin, 1965]{Lin1965}
Lin, S. (1965).
\newblock Computer solutions of the traveling salesman problem.
\newblock {\em The Bell System Technical Journal}, 44(10):2245--2269.

\bibitem[Luby et~al., 1993]{Luby1993}
Luby, M., Sinclair, A., and Zuckerman, D. (1993).
\newblock Optimal speedup of las vegas algorithms.
\newblock {\em Inf Process Lett}, 47(4):173--180.

\bibitem[Mitchell, 1998]{Mitchell1998}
Mitchell, M. (1998).
\newblock {\em An Introduction to Genetic Algorithms}.
\newblock MIT Press, Cambridge, MA.

\bibitem[Mitchell et~al., 1992]{Mitchell1992}
Mitchell, M., Forrest, S., and Holland, J. (1992).
\newblock The royal road for genetic algorithms: Fitness landscapes and ga
  performance.
\newblock In {\em ECAL}.

\bibitem[Mukhopadhyay et~al., 2014]{Mukhopadhyay2014}
Mukhopadhyay, A., Maulik, U., Bandyopadhyay, S., and Coello, C. A.~C. (2014).
\newblock A survey of multiobjective evolutionary algorithms for data mining:
  Part i.
\newblock {\em IEEE TEVC}, 18(1):4--19.

\bibitem[{National Academies}, 2019]{NAP25303}
{National Academies} (2019).
\newblock {\em Reproducibility and Replicability in Science}.
\newblock National Academies Press, Washington, DC.

\bibitem[Neri and Cotta, 2012]{Neri2012}
Neri, F. and Cotta, C. (2012).
\newblock Memetic algorithms and memetic computing optimization: A literature
  review.
\newblock {\em Swarm and Evolutionary Computation}, 2:1--14.

\bibitem[Nesmachnow et~al., 2015]{Nesmachnow2015}
Nesmachnow, S., Luna, F., and Alba, E. (2015).
\newblock An empirical time analysis of evolutionary algorithms as c programs.
\newblock {\em Softw Pract Exp}, 45(1):111--142.

\bibitem[Oliver et~al., 1987]{Oliver1987}
Oliver, I.~M., Smith, D.~J., and Holland, J. R.~C. (1987).
\newblock A study of permutation crossover operators on the traveling salesman
  problem.
\newblock In {\em Int Conf on Genetic Algorithms and Their Application}, pages
  224--230.

\bibitem[Petke et~al., 2018]{Petke2018}
Petke, J., Haraldsson, S.~O., Harman, M., Langdon, W.~B., White, D.~R., and
  Woodward, J.~R. (2018).
\newblock Genetic improvement of software: A comprehensive survey.
\newblock {\em IEEE TEVC}, 22(3):415--432.

\bibitem[Pizzuti, 2018]{Pizzuti2018}
Pizzuti, C. (2018).
\newblock Evolutionary computation for community detection in networks: A
  review.
\newblock {\em IEEE TEVC}, 22(3):464--483.

\bibitem[Ponsich et~al., 2013]{Ponsich2013}
Ponsich, A., Jaimes, A.~L., and Coello, C. A.~C. (2013).
\newblock A survey on multiobjective evolutionary algorithms for the solution
  of the portfolio optimization problem and other finance and economics
  applications.
\newblock {\em IEEE TEVC}, 17(3):321--344.

\bibitem[Prügel-Bennett, 2004]{PrugelBennett2004}
Prügel-Bennett, A. (2004).
\newblock When a genetic algorithm outperforms hill-climbing.
\newblock {\em TCS}, 320(1):135--153.

\bibitem[Scott and Luke, 2019]{Scott2019}
Scott, E.~O. and Luke, S. (2019).
\newblock Ecj at 20: Toward a general metaheuristics toolkit.
\newblock In {\em GECCO}, pages 1391--1398.

\bibitem[Selman and Gomes, 2006]{Selman2006}
Selman, B. and Gomes, C.~P. (2006).
\newblock Hill-climbing search.
\newblock In {\em Encyclopedia of Cognitive Science}, pages 333--336. Wiley.

\bibitem[Serpell and Smith, 2010]{Serpell2010}
Serpell, M. and Smith, J.~E. (2010).
\newblock Self-adaptation of mutation operator and probability for permutation
  representations in genetic algorithms.
\newblock {\em Evolutionary Computation}, 18(3):491--514.

\bibitem[Simson, 2019]{Simson2019}
Simson, J. (2019).
\newblock Lgp: A robust linear genetic programming implementation on the jvm
  using kotlin.
\newblock {\em JOSS}, 4(42):1337.

\bibitem[Snyk, 2024]{snyk}
Snyk (2024).
\newblock Snyk.
\newblock \url{https://snyk.io/}.

\bibitem[Sobania et~al., 2023]{Sobania2023}
Sobania, D., Schweim, D., and Rothlauf, F. (2023).
\newblock A comprehensive survey on program synthesis with evolutionary
  algorithms.
\newblock {\em IEEE TEVC}, 27(1):82--97.

\bibitem[{SpotBugs}, 2024]{spotbugs}
{SpotBugs} (2024).
\newblock {SpotBugs}: Find bugs in java programs.
\newblock \url{https://spotbugs.github.io/}.

\bibitem[Spotify, 2024]{spotify}
Spotify (2024).
\newblock fmt-maven-plugin: Opinionated maven plugin that formats your java
  code.
\newblock \url{https://github.com/spotify/fmt-maven-plugin}.

\bibitem[Starkweather et~al., 1991]{Starkweather1991}
Starkweather, T., McDaniel, S., Mathias, K., Whitley, D., and Whitley, C.
  (1991).
\newblock A comparison of genetic sequencing operators.
\newblock In {\em ICGA}, pages 69--76.

\bibitem[Syswerda, 1991]{Syswerda1991}
Syswerda, G. (1991).
\newblock Schedule optimization using genetic algorithms.
\newblock In {\em Handbook of Genetic Algorithms}. Van Nostrand Reinhold.

\bibitem[Szu and Hartley, 1987]{Szu1987}
Szu, H. and Hartley, R. (1987).
\newblock Nonconvex optimization by fast simulated annealing.
\newblock {\em Proceedings of the IEEE}, 75(11):1538--1540.

\bibitem[Tarkowski, 2023]{Tarkowski2023}
Tarkowski, T. (2023).
\newblock Quil\"{e}: C++ genetic algorithms scientific library.
\newblock {\em JOSS}, 8(82):4902.

\bibitem[Tayarani-N. et~al., 2015]{Tayarani2015}
Tayarani-N., M.-H., Yao, X., and Xu, H. (2015).
\newblock Meta-heuristic algorithms in car engine design: A literature survey.
\newblock {\em IEEE TEVC}, 19(5):609--629.

\bibitem[Valenzuela, 2001]{Valenzuela2001}
Valenzuela, C.~L. (2001).
\newblock A study of permutation operators for minimum span frequency
  assignment using an order based representation.
\newblock {\em J Heuristics}, 7(1):5--21.

\bibitem[\v{S}tefankovi\v{c} et~al., 2009]{Stefankovic2009}
\v{S}tefankovi\v{c}, D., Vempala, S., and Vigoda, E. (2009).
\newblock Adaptive simulated annealing: A near-optimal connection between
  sampling and counting.
\newblock {\em JACM}, 56(3):18:1--18:36.

\bibitem[Wan et~al., 2023]{Wan2023}
Wan, Y., Ma, A., He, W., and Zhong, Y. (2023).
\newblock Accurate multiobjective low-rank and sparse model for hyperspectral
  image denoising method.
\newblock {\em IEEE TEVC}, 27(1):37--51.

\bibitem[Whitley et~al., 1989]{Whitley1989}
Whitley, L.~D., Starkweather, T., and Fuquay, D. (1989).
\newblock Scheduling problems and traveling salesmen: The genetic edge
  recombination operator.
\newblock In {\em ICGA}, pages 133--140.

\bibitem[Xue et~al., 2016]{Xue2016}
Xue, B., Zhang, M., Browne, W.~N., and Yao, X. (2016).
\newblock A survey on evolutionary computation approaches to feature selection.
\newblock {\em IEEE TEVC}, 20(4):606--626.

\bibitem[Yazdani et~al., 2021]{Yazdani2021}
Yazdani, D., Cheng, R., Yazdani, D., Branke, J., Jin, Y., and Yao, X. (2021).
\newblock A survey of evolutionary continuous dynamic optimization over two
  decades—part a.
\newblock {\em IEEE TEVC}, 25(4):609--629.

\bibitem[Zhou et~al., 2021]{Zhou2021}
Zhou, X., Qin, A.~K., Gong, M., and Tan, K.~C. (2021).
\newblock A survey on evolutionary construction of deep neural networks.
\newblock {\em IEEE TEVC}, 25(5):894--912.

\end{thebibliography}

\end{document}